# NIV: Neural Ax*is* Variations for Variable Font Generation

**Nadav Benedek**
Reichman University, Two Delta
nadav.benedek@post.runi.ac.il

**Ariel Shamir**
Reichman University
arik@runi.ac.il

**Ohad Fried**
Reichman University
ofried@runi.ac.il

## Abstract

Variable fonts enable continuous variation of glyph geometry along semantic design axes such as weight, width, slant, and optical size. However, constructing a variable font from a static font remains a labor-intensive process requiring expert typographic design and manual specification of glyph variation data. We introduce NIV (Neural Axis Variations), a method that automatically converts a static font into a fully functional variable font. Given glyph outlines and a set of desired design axes, NIV predicts per-point displacements. The model operates directly on vector glyph geometry and employs a novel *Property Embedding* mechanism that captures interactions between multiple axes, enabling consistent multi-axis variation within a unified framework. We train NIV on a newly constructed dataset derived from variable Google Fonts, comprising over one million variation tuples. The resulting model generalizes across unseen code points, unseen font styles, high-complexity CJK glyphs, and even out-of-distribution handwriting inputs. The generated outputs are standard variable font files supporting continuous interpolation via existing rendering engines. To facilitate research, we release the dataset, the complete training and inference implementation, and trained models. Beyond typography, our approach demonstrates how structured geometric objects with continuous parametric variation can be synthesized using neural deformations.

## 1 Introduction

Variable fonts are an extension of the OpenType font format that allow continuous variation along one or more design axes within a single font file. These axes are defined by the type designer and can represent stylistic changes. Examples include weight, width, slant, and optical size (Appendix J), but designers are free to introduce custom axes for other transformations. Each axis defines a numerical range, for example weight from light to bold, and a specific font style is obtained by choosing a value on each axis. Instead of having separate font files for styles like Regular and Bold, a variable font encapsulates these in one font and enables smooth interpolation between styles. This offers flexibility in responsive design and typography. However, creating a variable font is a labor-intensive process traditionally done by expert designers. It requires designing multiple master glyph outlines and defining the `gvar` glyph variation data that specifies how glyph shapes change as a function of the axis values. Each interpolated instance corresponds to a particular combination of axis settings. Many existing fonts remain static, meaning single-instance, due to the effort required to add variations.

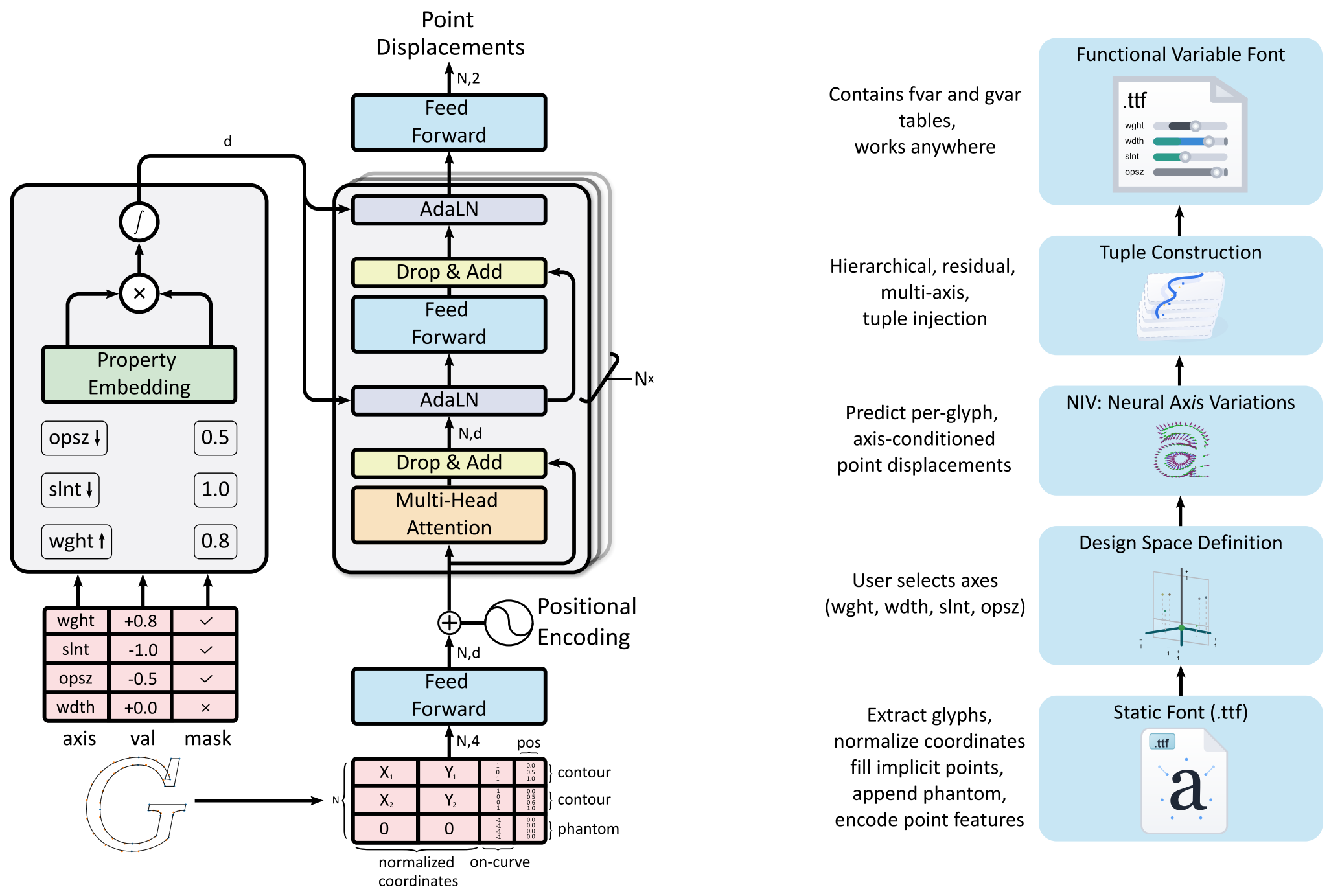

Figure 1: The NIV model and font generation method.

Table 1: Rendered pangram using three widely used, previously static fonts (Arial, Calibri, and Times New Roman), for which we generate, for the first time, corresponding variable fonts, shown under different axis configurations. Try selecting the text to verify that it is a real, functional font; since it is a genuine vector font, it can also be zoomed in arbitrarily without loss of quality.

| Configuration | Arial | Calibri | Times New Roman |
|---|---|---|---|
| default | The five boxing wizards jump quickly | The five boxing wizards jump quickly | The five boxing wizards jump quickly |
| wght<br>slnt | ***The five boxing wizards jump quickly*** | ***The five boxing wizards jump quickly*** | ***The five boxing wizards jump quickly*** |
| wght<br>slnt<br>wdth | The five boxing wizards jump quickly | The five boxing wizards jump quickly | The five boxing wizards jump quickly |
| wght<br>slnt<br>wdth | The five boxing wizards jump quickly | ***The five boxing wizards jump quickly*** | ***The five boxing wizards jump quickly*** |
| wght<br>slnt<br>wdth<br>opsz | ***The five boxing wizards jump quickly*** | ***The five boxing wizards jump quickly*** | ***The five boxing wizards jump quickly*** |
| wght<br>slnt<br>wdth<br>opsz | **The five boxing wizards jump quickly** | **The five boxing wizards jump quickly** | **The five boxing wizards jump quickly** |

In this paper, we introduce NIV (Neural Axis Variations)[1], a method for automatically generating a variable font from a static font. Given a TrueType font or a set of glyph outlines as input, the method produces a new font file that includes user-specified variation axes. For each axis and for combinations of axis values, the generated font contains the corresponding outline adjustments for every glyph. The model learns to predict outline deformations, represented as delta vectors on glyph control points, that correspond to movement along a design axis. By training on existing variable fonts, the model captures common patterns of shape change, such as stroke thickening for increased weight or proportion shifts for width expansion, and transfers this knowledge to new fonts and to Unicode characters that were not observed during training. To our knowledge, NIV is the first method to generate a variable font file with continuous variations from a single static font. Prior neural glyph generation work has largely focused on producing static outputs, either raster images or vector outlines for individual glyphs, which do not result in a functioning font and do not support adjustable axes. In contrast, our method outputs a standard OpenType variable font (TTF), enabling immediate use in browsers or design software via a continuous axis slider. This PDF itself serves as a demonstration, as it uses variable fonts generated by our method throughout the document. The rendered text is selectable and reflects continuous axis variations.

[1]This document should be viewed using a standards-compliant desktop PDF viewer (e.g., Acrobat, Foxit, or Preview), as certain browser-based PDF renderers do not yet fully support OpenType variable fonts.

## 2 Variable Font Structure

A variable font extends the TrueType/OpenType format by defining one or more design axes of the font variation, forming a design space. A position in the design space affects the shapes of the font glyphs. Several font tables support this functionality. The `fvar` (font variations) table defines the axes themselves. For each design axis, `fvar` specifies parameters like the axis minimum, default, and maximum values, and often some specific named instances (such as "Bold" at `wght`=700 if default is 400). In our case, when creating a new axis, we add an entry to `fvar` and define its range. We typically set the default value to 0 (the original font's design), and allow variation from -1 to +1. The heart of the variability functionality lies in the `gvar` (glyph variations) table. The `gvar` table contains *tuple* variation data for each glyph.

A *tuple* represents a specific adjustment (delta) to a glyph's outline at a given range in the design space, defined by axis coordinates at the edges. For example, a weight-axis *tuple* for the letter "a" may specify how each outline point moves from its original position (`wght`=0) to create the bold extremes (-1 or +1). Each *tuple* consists of two components: (1) a range (for each axis) defined by minimum, peak, and maximum values that specifies where the tuple is active; and (2) a set of deltas, namely for each point in the glyph outline (including off-curve control points and *phantom points*), a displacement applied at that coordinate. A formal mathematical description of the interpolation mechanism is provided in Appendix A.

## 3 Method

Given a non-variable (static) font and a set of desired design axes, our method generates a corresponding variable font that supports these axes automatically. We use a two-stage approach. In the first stage, we train the NIV model to predict how an individual glyph outline deforms, conditioned on the variation axis values. Figure 1 illustrates the architecture. For an input glyph, represented as an ordered sequence of vector outline points, and a position in the design space (e.g. [`wght`=0.5, `slnt`=0.0, `wdth`=-1.0]), the model predicts the corresponding sequence of displacement vectors $(\Delta x, \Delta y)$ for each glyph's outline point. In the second stage, we use this model to build each *tuple* in the `gvar` table for each glyph for constructing the variable font.

### 3.1 Architecture

This section describes the NIV model's input representation, internals, and training.

#### 3.1.1 Input

Each glyph outline is extracted from the font's `glyf` table and represented as a sequence of points. TrueType glyphs consist of one or more contours, each represented as a closed sequence of on-curve and off-curve points interpreted as a piecewise quadratic Bézier spline, where segments may be either curved or straight. The glyph is encoded as an ordered sequence of points, where each point is associated with a set of features: its normalized $(x, y)$ coordinates, a scalar on-curve indicator distinguishing on-curve points from off-curve control points, and contour-related information that enables the model to identify contour boundaries and the relative position of points within each contour. Let $N_g$ denote the total number of control points in the glyph. We append the four phantom points, so the total sequence length is $N = N_g + 4$, as shown in Figure 1. All glyph coordinates are normalized to a common reference frame by mapping the font's units-per-em (UPM) square to a fixed range. Specifically, outline coordinates $(x, y)$ are scaled by the inverse of the font's UPM value (typically 1000 or 2048) and shifted so that positions lie approximately in the interval $[-1, 1]$. The same scaling is applied to the displacement vectors $(\Delta x, \Delta y)$, ensuring consistency between inputs and regression targets. This normalization improves training stability and facilitates generalization across fonts with different absolute scales. The on-curve/off-curve indicator is encoded as a scalar feature, with values $1$ for on-curve points, $0$ for off-curve control points, and $-1$ for phantom points. Points are assigned a normalized contour position parameter in the range $[0, 1]$, evenly spaced according to their order along the contour, allowing the model to capture relative position within each contour while remaining invariant to contour length.

Along with the glyph point sequence, the model receives a set of target variation axes together with their desired values. Each axis is specified by an axis identifier and a normalized scalar value in the range $[-1, 1]$, indicating the strength and direction of variation along that axis. The axis identifier is a lookup key selecting the learned embedding pair associated with a given axis, analogous to a token ID in a language model. Multiple axes may be provided simultaneously, allowing the model to condition on a joint location in the multi-dimensional design space. A binary mask indicates which axis slots are active, enabling a single model to handle a variable number of axes per example.

#### 3.1.2 Model

The proposed method introduces a novel sequence-to-sequence geometric model that maps an ordered sequence of glyph point features to a sequence of outline displacement vectors. The model operates directly on variable-length point sequences and captures geometric dependencies between points. This is essential for glyph outlines, whose complexity and number of points vary widely across characters and scripts. Each input point is first mapped from its low-dimensional geometric description to a $d$-dimensional latent representation using a learned point-wise feed-forward projection. Positional encoding is then added to the resulting sequence to preserve the serialization order induced by the contour structure.

The core of the model consists of a stack of identical interaction blocks, each operating on the full point sequence and producing an updated sequence of the same length. Within each block, point representations are jointly updated via multi-head self-attention [Vaswani et al., 2017] followed by feed-forward transformations, with residual connections and dropout applied throughout. The feed-forward sub-network within each encoder layer consists of two linear transformations with a GELU [Hendrycks and Gimpel, 2016] activation and dropout in between. The first expands the latent dimension by a factor of 4, and the second contracts to the original dimension. All blocks process the entire sequence simultaneously rather than autoregressively, reflecting the fact that glyph outlines are geometric structures rather than temporally ordered signals.

#### 3.1.3 Property Embedding

A central contribution of this work is a novel *Property Embedding* mechanism for conditioning the model on a set of axes, each specified by an axis identifier, a normalized scalar value in $[-1, 1]$, and a binary mask indicating whether axis conditioning is active. This representation supports a variable number of axes per example. Given $A$ possible axes, the embedding space contains two learned vectors (positive and negative) for every axis, resulting in $2A$ embeddings. The corresponding embedding is weighted by the product of the absolute axis values, and all such weighted embeddings are summed to produce a single conditioning vector. We further test whether joint multi-axis conditioning is necessary in Appendix H. The resulting conditioning vector is normalized to maintain a consistent scale across different numbers of active axes. Property Embedding conditions the network through Adaptive Layer Normalization (AdaLN, Peebles and Xie [2023]) at every interaction block. Concretely, the axis-conditioning vector is used to generate the scale and shift parameters of each layer normalization operation. This allows the model to modulate the internal representations at all depths of the network, enabling fine-grained, geometry-aware control of the deformation process while maintaining stable training dynamics. After the final interaction block, a point-wise regression head maps each $d$-dimensional latent point representation to a two-dimensional displacement vector $(\Delta x, \Delta y)$. Implementation details are provided in Appendix F, and ablations in Appendix I.

#### 3.1.4 Training

The model is trained in a supervised manner using ground-truth outline differences computed between the default glyph and a deformed glyph at the specified axis configuration. A mean squared error loss is applied independently to all points. The model operates purely on geometry and does not receive explicit glyph identity or Unicode information, enabling it to generalize deformation patterns to unseen characters based solely on shape. To train the NIV model, we use a dataset constructed from existing variable fonts available through Google Fonts [2010], where glyphs include designed variations across design axes. Each training example input consists of a static glyph outline from a specific font and axis coordinates extracted from the peak axis values of a specific tuple. The target label is the glyph's deformation under those axis settings, computed by combining all corresponding variation tuples as specified by the variable font standard. There are multiple options for the train/test split strategy, each designed to evaluate a different aspect of the model's generalization:

**Unicode split:** Glyphs are divided by their Unicode code points. A particular Unicode character appears exclusively in either the training or test set. This split evaluates the model's ability to generalize deformation patterns to unseen code points based solely on geometric structure rather than character identity.

**Font split:** Entire fonts are assigned to either train or test. This means all glyphs from a single font will be in the same subset, enabling evaluation of generalization across styles of fonts. In this setting, the same Unicode code points may appear in both splits, so generalization is measured across unseen font styles rather than unseen characters.

A combined Unicode and font split is possible, but would result in substantial data waste, since glyphs belonging to test fonts that share Unicode code points with the training set would need to be discarded.

### 3.2 Font Generation

In the second stage, we apply the trained model to construct a variable font from a non-variable font. First, we select the axes we want to define in the design space and build the `fvar` table. For each glyph, the model predicts the outline deltas for each position in the design space. We start with sub-spaces of one axis at a time, and build up to higher-order sub-spaces, and for each one we sample the extreme coordinates for each axis. This allows us to inject `gvar` tuples progressively by order, first adding first-order tuples for individual axes, followed by higher-order tuples for pairs, triples, and up to a predefined maximum order specified by the user (typically capped at four axes). Higher-order tuples are written as *residual* corrections: for a target multi-axis coordinate, we first compute the displacement already explained by previously inserted lower-order tuples using the OpenType tuple-weighting rule (A.4), and subtract this accumulated contribution from the model's predicted displacement before writing the new tuple. As a result, first-order tuples capture single-axis effects, while higher-order tuples encode only the additional interaction terms required for accurate multi-axis interpolation. This mirrors the hierarchical structure used in professional variable fonts, while learning all tuple deltas from data. For the experiments in this paper, we produce variable fonts using the four most common axes in the dataset (`wght`, `wdth`, `opsz`, and `slnt`) as summarized in Table 18.

### 3.3 Invariance to Axis and Tuple Order

Neither the order in which axes are given to the model, nor the order in which tuples are constructed or stored, affects the final variable font.

**Axis order into the model.** The Property Embedding (Section 3.1.3) forms its conditioning vector by summing a weighted embedding for each active axis. Since summation is commutative, permuting the input axes leaves the conditioning vector, and hence the model's prediction, unchanged.

**Tuple order in the compiled font file.** The OpenType format itself is order-invariant: a glyph's outline at any coordinate is a flat sum over all stored tuples (6). Storage order therefore cannot affect rendering.

**Order of construction.** We build tuples hierarchically by order: first single-axis tuples, then pairs, then triples, and so on, with each higher-order tuple written as a residual after subtracting the contribution of lower-order tuples. Every tuple we construct sits at an extremum ($+1$ or $-1$) for its own axes and at the default value ($0$) for every other axis. The tent function (3) is exactly zero at this default value, so two sibling tuples of the same order (e.g., `wght+wdth` and `wght+opsz`) never contribute to each other's residual: each is zero wherever the other is being evaluated. Consequently, the residual for any tuple depends only on strictly lower-order tuples, never on same-order siblings, so the order in which same-order tuples are constructed cannot change the result.

Together, these three arguments show that our construction procedure is fully invariant to axis and tuple order.

# 4 Public Dataset

To facilitate reproducibility and future research, we release a newly constructed dataset derived from the public variable Google Fonts, together with the full preprocessing pipeline. The dataset includes per-glyph outline geometry and fully densified per-point variation deltas extracted from the `gvar` tables, normalized to a unified coordinate system. The dataset is split into training and test sets and released in a structured XML format. A detailed description of parsing, densification of sparse deltas, phantom-point handling, axis filtering, and serialization is provided in Appendix L.

# 5 Experiments

## 5.1 Evaluation Metric

We evaluate NIV using the root mean squared error (RMSE) over all test examples. Each test example consists of an input and a label. The input comprises the base glyph outline, a sequence of control point coordinates, each scaled by $2/\text{UPM}$ and shifted to lie approximately in $[-1, 1]$, together with an on-curve/off-curve indicator per point, and an axis coordinate indicating a position along the variation axis. The label is the total per-point displacement at that axis coordinate, computed by summing the displacements from all variation tuples defined for the glyph, each weighted by its tent-function evaluated at the given axis coordinate, and scaled by $2/\text{UPM}$ to match the normalized coordinate space. Test examples are generated by iterating over every font in the test set, then over every glyph in each font, and then over every variation tuple defined for that glyph. Each tuple specifies an axis peak coordinate, which becomes the conditioning input. The label, however, is not the displacement stored in that single tuple; rather, it is the total displacement obtained by accumulating contributions from *all* tuples of the glyph, weighted according to the tent function defined by each tuple's coordinate specification, evaluated at the current axis coordinate. During evaluation, the model receives the input and predicts a displacement vector for each control point. For each test example, we compute the squared Euclidean distance between the predicted and ground-truth displacement at every control point, including the four phantom points, and average these values across all points and all test examples.

## 5.2 Static-to-Variable Font Generation

To evaluate NIV's ability to generalize to unseen font styles, we tested the method on static fonts that were not included in the training dataset. Each complete `.ttf` file was processed by the model pipeline, which crafted a novel variable font. Table 1 presents qualitative results for the generated variable fonts, rendered at different coordinates within the design axes. Additional examples are provided in Appendix B. These examples demonstrate the model's ability to synthesize stylistically coherent and smooth interpolations across multiple font styles and axis configurations. Other font-level metrics such as kerning, spacing, and hinting are outside the scope of this paper. However, they can be readily injected into the generated variable font post-hoc to improve visual quality. Learning these additional aspects automatically is a promising direction for future research.

## 5.3 Numerical Evaluation under Different Split Strategies

To quantify generalization under different notions of data separation, Table 2 reports the test-set root mean squared error (RMSE) for the Unicode split and font split strategies. The Unicode split evaluates the model's ability to generalize deformation patterns to previously unseen code points, while the font split measures generalization to entirely unseen font styles. For each split strategy, we train the model with different random seeds and report the mean and standard deviation of the final test RMSE. Consistent with expectations, the test RMSE for the Unicode-based split is slightly higher than for the font-based split. In the former setting, the model must generalize to previously unseen characters, whereas in the latter it encounters familiar characters and is required only to generalize across novel stylistic variations. Importantly, for a new style to remain legible to human readers, it must preserve structural regularities shared with existing styles, which may facilitate generalization. In contrast, unseen characters introduce genuinely new structural configurations, making the task inherently more challenging.

Table 2: Test-set root mean squared error (RMSE) under different split strategies of our model predictions (mean $\pm$ $1\sigma$ standard deviation over three consecutive seeds), compared to a null-baseline in which all delta predictions are zero. RMSE is calculated in normalized delta units, therefore an RMSE of 1.0 corresponds to 1024 font units, equal to half of the em box (for UPM=2048).

| **Train/Test split** | **Purpose** | **NIV Test RMSE ↓** | **Null Baseline RMSE ↓** |
|---|---|---|---|
| Unicode split | Generalization to unseen code points | $\mathbf{0.05354 \pm 0.00023}$ | 0.13503 |
| Font split | Generalization across font styles | $\mathbf{0.05205 \pm 0.00088}$ | 0.14433 |

### 5.4 Reconstructing Variable Fonts in the Test Set

One way to visually evaluate the quality of the NIV model is to reconstruct the variable font files in the test split. In this experiment, we take variable font files, strip out all variations, effectively converting them into static font files, and then apply our method to predict the variations. Table 3 shows a side-by-side comparison between the Commissioner, Saira-Italic and RadioCanada variable fonts and the reconstruction produced by our model. Overall, high fidelity is observed, and the results show that our method successfully predicts the font variations, with a few noticeable imperfections. When analyzing the comparison, kerning (the spacing between characters) should be ignored, as it is not predicted by the model and is outside the scope of this paper.

### 5.5 Robustness on High-Complexity CJK Fonts

To evaluate the robustness of our variable font generation model under challenging linguistic and typographic conditions, we assessed its performance on CJK fonts. Specifically, we selected Meiryo and PingFang, static Japanese and Chinese fonts that are not part of the training data. Designing fonts for CJK languages is significantly more demanding than for Western scripts due to the vast number of required glyphs, typically in the thousands. This complexity becomes even more pronounced when creating variable fonts, which traditionally require extensive manual work to define consistent and meaningful deformations across all glyphs. Table 4 and Appendix E present qualitative results demonstrating how the generated variable versions perform when applied to Chinese, Japanese and Korean sentences.

### 5.6 Out-of-Distribution Generalization: Handwriting

To test the generalization capability of our method on truly unseen and fundamentally novel inputs, we created a static font from a person's handwriting sample, a style that bears no structural resemblance to any character shapes present in the training data. The input consisted only of individual glyph outlines without any kerning or font-level metadata. We then applied NIV to generate a variable font. The resulting variable font allows interpolation along these axes, even though the original input was a single static design. This represents a highly challenging setting for the model, as it must infer axis-driven deformations for glyph shapes it has never encountered during training, demonstrating strong generalization beyond the training distribution. Table 5 presents a pangram rendered in the generated variable font under different axis settings. Additional examples are in Appendix C.

Table 3: Reconstructing variable fonts in the test set. Rendered pangram using variable fonts (from test set) and the corresponding NIV-generated variable fonts (reconstruction) under matched axis configurations.

| Configuration | Original | NIV |
|---|---|---|
| **Commissioner** | | |
| slnt | The five boxing wizards jump quickly | The five boxing wizards jump quickly |
| wght | The five boxing wizards jump quickly | The five boxing wizards jump quickly |
| wght<br>slnt | The five boxing wizards jump quickly | The five boxing wizards jump quickly |
| **Saira-Italic** | | |
| wght | The five boxing wizards jump quickly | The five boxing wizards jump quickly |
| wght | The five boxing wizards jump quickly | The five boxing wizards jump quickly |
| wdth | The five boxing wizards jump quickly | The five boxing wizards jump quickly |
| wght<br>wdth | The five boxing wizards jump quickly | The five boxing wizards jump quickly |
| **RadioCanada** | | |
| wght | The five boxing wizards jump quickly | The five boxing wizards jump quickly |
| wdth | The five boxing wizards jump quickly | The five boxing wizards jump quickly |
| wght<br>wdth | The five boxing wizards jump quickly | The five boxing wizards jump quickly |

Table 4: Rendered CJK text using novel variable fonts automatically generated from static Meiryo (Japanese) and PingFang (Chinese) typefaces using NIV.

| Axis Configuration | Rendered Text |
|---|---|
| **Japanese (Meiryo)** | |
| original | 茶色のキツネは怠け者の犬を飛び越えた |
| wght slnt wdth | 茶色のキツネは怠け者の犬を飛び越えた |
| wght slnt wdth | 茶色のキツネは怠け者の犬を飛び越えた |
| wght slnt wdth | 茶色のキツネは怠け者の犬を飛び越えた |
| wght slnt wdth | 茶色のキツネは怠け者の犬を飛び越えた |
| wght slnt wdth | 茶色のキツネは怠け者の犬を飛び越えた |
| **Chinese (PingFang)** | |
| original | 敏捷的棕色狐狸跳過了懶狗 |
| wght slnt wdth | 敏捷的棕色狐狸跳過了懶狗 |
| wght slnt wdth | 敏捷的棕色狐狸跳過了懶狗 |
| wght slnt wdth | 敏捷的棕色狐狸跳過了懶狗 |
| wght slnt wdth | 敏捷的棕色狐狸跳過了懶狗 |
| wght slnt wdth | 敏捷的棕色狐狸跳過了懶狗 |

Table 5: Rendered pangram using the generated variable font from a handwriting sample under different axis settings. Try selecting the text to verify that it is a real font and not a static image.

| Axis Configuration | Rendered Text |
|---|---|
| wght slnt wdth opsz | The five boxing wizards jump quickly |
| wght slnt wdth opsz | The five boxing wizards jump quickly |
| wght slnt wdth opsz | The five boxing wizards jump quickly |
| wght slnt wdth opsz | The five boxing wizards jump quickly |

### 5.7 Comparison to Rule-Based Geometric Baselines

Generating axis-based variations without learning requires hand-crafted geometric heuristics. Since there is no general classical algorithm for constructing multi-axis variable fonts, we implement simple rule-based geometric baselines to illustrate the limitations of such approaches. These heuristics are typically axis-specific, do not generalize across multiple axes, and cannot capture semantic deformation behavior. In contrast, NIV learns a unified deformation space from data, enabling variation across a range of axes. As a representative example of why such geometric approaches fall short in practice, consider the difficulty of constructing a width-axis variation using simple transformations. A naive approach might apply a horizontal scaling to increase glyph width. However, this also inadvertently thickens vertical stems, an undesirable artifact, as width-axis variation is expected to preserve stem thickness while adjusting only the overall horizontal proportions. Figure 3a contrasts this naive transformation with the output of NIV: while the geometric baseline alters stem thickness, NIV preserves structural integrity by selectively deforming only the appropriate components. See Appendix G for further evidence of the limitations of such rule-based approaches.

## 6 Related Work

**Neural Raster Glyph Generation** Early work [Baluja, 2017, Azadi et al., 2018] focused on generating glyphs as raster (pixel-based) images. Subsequent approaches formulate glyph synthesis as an image-to-image translation or few-shot learning problem, generating individual glyphs in pixel space rather than vector outlines, and lacking mechanisms for variable fonts. Xie et al. [2021] employ convolutional networks to transfer visual style from a small set of reference glyph images to unseen characters. Other raster glyph generation methods focus on synthesizing glyphs as raster images from a small number of reference examples by disentangling content and style representations. Gao et al. [2019] and Park et al. [2021a] formulate glyph generation as a few-shot learning problem and learn to transfer stylistic features across characters. Cha et al. [2020] proposed a dual-memory architecture to enable compositional glyph generation from limited supervision, while Park et al. [2021b] extend few-shot synthesis using a mixture-of-experts framework to better capture diverse font styles.

**Static Vector Glyph Generation** Generating vector glyphs is more challenging due to the discrete and structured representation of font outlines. Prior to neural approaches, example-based methods generated vector glyphs by transferring stroke or component geometry from reference characters, including calligraphy synthesis Xu et al. [2005] and example-based font generation via component matching Suveeranont and Igarashi [2010]. One of the first neural approaches for vector graphics was Lopes et al. [2019], which learns a representation of SVG graphics by encoding rasterized images and decoding vector command sequences. This line of work was later extended by Carlier et al. [2020], a hierarchical VAE designed for scalable vector graphics, demonstrated primarily on icon-like SVGs. Wang and Lian [2021] synthesize font glyphs directly as Bézier curves, and adopt a dual-modality representation, combining raster glyph images with their corresponding vector outlines as input, and generates new glyph outlines in the same style. Wang et al. [2023] replaces recurrent networks with Transformer architectures to better model long outline sequences and introduces a refinement module to reduce geometric artifacts. These approaches improve vector glyph quality, producing more faithful outlines by better aligning generated curves to target shapes. Recently, diffusion models have been explored for glyph generation. Thamizharasan et al. [2024] is a two-stage diffusion approach: a first stage generates a low-resolution raster image with rough shape and key points, and a second stage generates the precise vector outline conditioned on the raster. Beyond font-specific outline synthesis, recent work has also studied general SVG and vector-pattern generation, including text-guided SVG synthesis with diffusion [Xing et al., 2024], exemplar-based synthesis of structured vector textures [Tu et al., 2022], and multimodal generation of SVG code from images and text [Rodriguez et al., 2025]. However, these methods still require multiple reference glyphs to capture a glyph style, generate glyphs independently one at a time, or operate on general vector graphics rather than font files. They do not produce a unified font file and are not designed to model continuous axes of variation, yielding only static glyph instances or standalone vector graphics rather than variable fonts.

## 7 Limitations and Future Work

While NIV produces high-quality variable fonts in most cases, several limitations remain. First, the method focuses primarily on geometric outline deformation. Although the model predicts displacements for all glyph points, including the four phantom points used to encode advance width and side bearings, it does not explicitly predict typographic components such as kerning pairs, global spacing rules, hinting instructions, or advanced OpenType layout tables. Extending the model to jointly predict glyph geometry, spacing behavior, and font-level layout tables is a promising direction for future work. Second, the training dataset is constructed solely from variable fonts available in Google Fonts. Expanding the dataset to include additional public or licensed variable font sources could increase stylistic diversity and improve generalization. Third, while our experiments focus on typographic glyphs, the proposed axis-conditioned deformation framework may extend to structured vector graphics beyond the font domain, suggesting a promising direction for future research.

## 8 Conclusion

We presented NIV, a neural method for automatically generating variable fonts from a single static font. By learning axis-conditioned geometric displacements directly over glyph control points, NIV produces functional OpenType variable fonts that support continuous interpolation along semantic design axes. Our contribution is a conditional geometric regression framework that respects contour structure, predicts continuous deformations, and explicitly represents interactions among multiple continuous properties, realized in part through Property Embedding, which models higher-order interactions between multiple axes in a permutation-invariant manner. Experiments demonstrate strong generalization across unseen Unicode characters, unseen font styles, high-complexity CJK scripts, and out-of-distribution handwriting inputs. NIV produces standard-compliant font files that can be immediately used in existing browsers and design tools. In many cases, the generated variable fonts require little to no manual adjustment; when refinement is needed, NIV reduces the amount of expert intervention compared to constructing variable fonts from scratch, thereby saving the majority of manual engineering effort. More broadly, NIV suggests a general paradigm for learning structured parametric variation in vector graphics and other geometric domains.

## Acknowledgments and Disclosure of Funding

This work was partially supported by Israel Science Foundation (grant no. 1427/25 and 1574/21). We thank Itzik Malkiel, Limor Benedek, Victoria Kaufman, and Oded Onn.

## Appendix Table of Contents

# A Formal Variable Font Structure

## A.1 Fonts and Glyphs

A *font* defines a mapping from characters to geometric shapes together with global layout and rendering information. At the geometric level, a font consists of a finite collection of *glyphs*, each representing the outline of a single character or symbol.

Formally, let $\mathcal{G}$ denote the set of glyphs in a font. Each glyph $g \in \mathcal{G}$ is represented by a sequence of control points forming one or more closed contours. We denote the default (non-variable) outline of glyph $g$ as

$$P_g^{(0)} = \{(x_i, y_i)\}_{i=1}^{N_g} \subset \mathbb{R}^2 \tag{1}$$

where $N_g$ is the total number of control points, including on-curve points and off-curve quadratic Bézier control points in glyph $g$. The superscript $(0)$ denotes the default glyph outline at the origin of the design space, corresponding to the font's unmodified (pre-variation) design. Figure 2 shows a representative example of such a default outline.

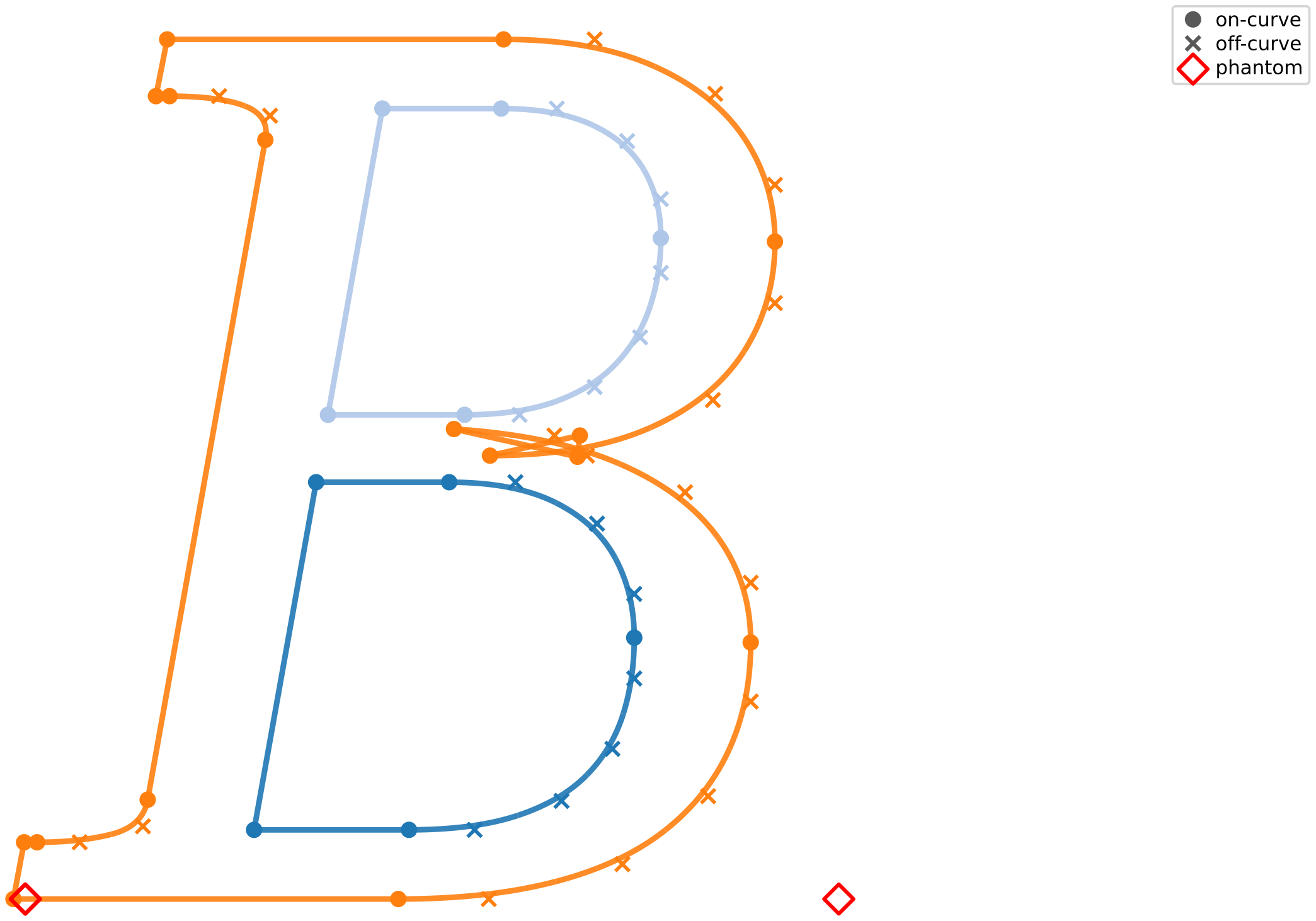

Figure 2: A glyph in RobotoSerif-Italic, rendered from its default (non-variable) outline $P_g^{(0)}$. Each contour is drawn in a different color, with on-curve points shown as filled circles and off-curve quadratic Bézier control points shown as crosses. A straight line segment is drawn between two consecutive on-curve points, whereas a quadratic Bézier curve is drawn between two on-curve points whenever a single off-curve point lies between them, using the off-curve point as the control point. When two or more off-curve points occur consecutively, as permitted by the TrueType outline format, an implied on-curve point is inserted at the midpoint of each consecutive off-curve pair, splitting the segment into multiple quadratic Bézier curves; these implied points are not explicitly stored in the font and are not marked in the figure. The four red diamonds are the glyph's phantom points, which encode its horizontal and vertical metrics. Three of them coincide with the origin $(0, 0)$: the left phantom point is at $(0, 0)$, while the top and bottom phantom points default to $(0, 0)$ because RobotoSerif-Italic has no vertical metrics table (`vmtx`), i.e., it does not support vertical text layout. In this example, only the right phantom point carries information, encoding the glyph's horizontal advance width.

### A.2 Design Axes and Design Space

A *variable font* augments a static font by introducing continuous *design axes* that parametrize families of related glyph shapes. Examples include weight (`wght`), width (`wdth`), slant (`slnt`), and optical size (`opsz`).

Let $A$ denote the number of axes. Each axis $a \in \{1, \ldots, A\}$ is defined by a scalar parameter with a user-facing range and a default value, as specified in the `fvar` table. User-facing axis values are mapped to an internal normalized coordinate system, optionally via a piecewise-linear remapping defined by the `avar` table.

Internally, axis values are represented as

$$z_a \in [-1, 1]$$

where $z_a = 0$ corresponds to the default design, and $z_a = \pm 1$ denote the axis extremes. A point in the *design space* is therefore given by the vector

$$z = (z_1, \ldots, z_A) \in [-1, 1]^A$$

### A.3 Glyph Variations and Tuples

Glyph variation is defined relative to the default outline. Rather than storing complete outlines for all axis configurations, a variable font stores in its `gvar` table a finite set of *tuple variations*, each describing how a glyph's outline changes over a localized region of the design space.

For each glyph $g \in \mathcal{G}$, the font designer defines $T_g$ tuple variations, indexed by $t = 1, \ldots, T_g$. Each tuple $t$ jointly specifies: (i) a region of the $A$-dimensional design space in which it is active, and (ii) a set of point-wise displacements applied to the glyph outline. Formally, a tuple is represented by a displacement field

$$\Delta_{g,t} = \{(\Delta x_i^{(t)}, \Delta y_i^{(t)})\}_{i=1}^{N_g+4}, \tag{2}$$

where $N_g$ is the number of on-curve points and off-curve quadratic Bézier control points explicitly stored in the `glyf` table, and the additional four points are *phantom points* implicitly defined by the TrueType specification to encode glyph metrics such as advance width and side bearings (illustrated in Figure 2). Although phantom points are not rendered and do not belong to any contour, they participate in variation and interpolation in the same manner as geometric points.

Different tuples typically correspond to different axis configurations. For example, one tuple may encode variation along the `wght` axis at its positive extreme (`wght` $= +1.0$), another may encode the negative extreme (`wght` $= -1.0$), while additional tuples may encode variation along other axes such as `slnt`. More generally, tuples may represent joint variations across multiple axes, for instance a tuple peaking at `wght` $= +1.0$ and `slnt` $= -0.5$, capturing higher-order interactions between axes.

### A.4 Tuple Support and Axis Regions

Each tuple $t$ defines, for each axis $a$, a one-dimensional support region that specifies where the tuple is active along that axis. Let $v_a \in [-1, 1]$ denote the normalized value (coordinate) of axis $a$. The support of tuple $t$ on axis $a$ is parameterized by a triple $(\ell_{t,a}, p_{t,a}, r_{t,a})$, corresponding to the left boundary, peak, and right boundary of the region, with $\ell_{t,a} \leq p_{t,a} \leq r_{t,a}$.

We define a one-dimensional *tent function*

$$\text{tent}(v; \ell, p, r) \ = \begin{cases} 0, & v \leq \ell \text{ or } v \geq r, \\ \dfrac{v - \ell}{p - \ell}, & \ell < v < p, \\ \dfrac{r - v}{r - p}, & p \leq v < r. \end{cases} \tag{3}$$

which rises linearly from zero at $\ell$ to one at $p$ and falls linearly back to zero at $r$.

The per-axis weight of tuple $t$ on axis $a$ is then given by

$$w_{t,a}(v_a) = \text{tent}(v_a; \ell_{t,a}, p_{t,a}, r_{t,a}), \tag{4}$$

with the convention that if tuple $t$ does not constrain axis $a$, then $w_{t,a}(v_a) = 1$.

In the dataset used in this paper, all tuple support regions follow the default OpenType "tent" behavior, meaning that only peak coordinates are explicitly stored. According to the OpenType specification, when only a peak value is provided, the support region is implicitly defined: for positive peak values the support extends from the default (left=0) to the peak (right=peak), and for negative peak values from the peak (left=peak) to the default (right=0). This produces the standard triangular (tent) influence function without explicitly storing start and end coordinates.

### A.5 Multi-Axis Interpolation

The overall weight of tuple $t$ at design-space configuration $v = (v_1, \dots, v_A)$ is obtained by combining the per-axis weights multiplicatively:

$$\lambda_{g,t}(v) = \prod_{a=1}^{A} w_{t,a}(v_a). \tag{5}$$

The glyph outline at axis configuration $v$ is then given by

$$P_g(v) = P_g^{(0)} + \sum_{t=1}^{T_g} \lambda_{g,t}(v)\, \Delta_{g,t}. \tag{6}$$

This formulation defines a continuous, piecewise-linear mapping from the design space to glyph geometry.

## B Additional Static-to-Variable Font Generation

To further illustrate the behavior of NIV on stylistically distinct inputs, we present results on *Brush Script*, a cursive typeface that differs substantially from the predominantly geometric and serif/sans-serif fonts in the Google Fonts training corpus. Brush Script is not included in the Google Fonts dataset and is distributed under licensing terms that prohibit redistribution of the font file. For this reason, we present rendered outputs rather than embedding or distributing the generated variable font itself. Table 6 shows the generated variations for Brush Script under different axis configurations. Despite the strong cursive characteristics and stroke-level dependencies of this typeface, NIV produces coherent multi-axis deformations while preserving the global stylistic identity of the font.

Table 7 presents variations generated by our model for the Gothic typeface UnifrakturMaguntia, which was previously available only as a static font. Despite the distinctive stylistic characteristics of Gothic letterforms, the model produces coherent and visually consistent variations across axes.

Table 8 presents variations generated for the symbolic typeface Wingdings, which differs fundamentally from alphabetic fonts in that glyphs correspond to pictographic symbols rather than letters. Despite the absence of conventional stroke structure and contour semantics, the model produces consistent multi-axis deformations while preserving the recognizability and geometric integrity of the original symbols. This suggests that NIV may extend beyond the font domain and could potentially be applied to more general vector graphics, which we view as a promising direction for future work.

## C Additional Handwriting Experiments

Additional handwriting samples were collected and converted into a static font and then to a variable font using NIV. The resulting variable fonts were then rendered with different axis settings to demonstrate the range of variations that can be achieved. The rendered pangrams for these additional handwriting samples are shown in Table 9.

## D Optical Size Generalization

Table 10 compare our model predictions for BodoniModaSC variable font from the test set with the variations chosen by its original designer, at various optical sizes.

The qualitative comparison demonstrates that the model captures the expected structural changes across optical sizes, including contrast adaptation and detail simplification, while preserving the characteristic style of BodoniModaSC, with slight inaccuracies when the stroke thickness becomes hairline thin.

## E CJK Fonts: Korean

In addition to Japanese and Chinese, we evaluate NIV on Korean by converting the static (non-variable) typeface Nanum Myeongjo into a variable font, as shown in Table 11.

Table 6: Generated variations for the static Brush Script font under different axis configurations. Since the original font is subject to licensing restrictions, we present rendered outputs rather than embedding the generated variable font file. To our knowledge, this is the first time a variable version of Brush Script has been automatically generated using a neural network. The resulting variations exhibit consistent multi-axis behavior and preserve the characteristic cursive structure of the original design, demonstrating both geometric accuracy and stylistic coherence.

| Axes | Rendered Text |
| --- | --- |
| wght<br>slnt<br>wdth<br>opsz | The quick brown fox jumps over the lazy dog |
| wght<br>slnt<br>wdth<br>opsz | The quick brown fox jumps over the lazy dog |
| wght<br>slnt<br>wdth<br>opsz | The quick brown fox jumps over the lazy dog |
| wght<br>slnt<br>wdth<br>opsz | The quick brown fox jumps over the lazy dog |
| wght<br>slnt<br>wdth<br>opsz | The quick brown fox jumps over the lazy dog |
| wght<br>slnt<br>wdth<br>opsz | The quick brown fox jumps over the lazy dog |

## F Training Setup and Implementation Details

This section and Table 12 summarize the architectural configuration, optimization hyperparameters, data splits, and compute resources used in all experiments. Unless otherwise stated, the same configuration is used across all split strategies reported in the main paper.

The latent dimensionality is set to $d = 256$, with 6 interaction blocks and 8 attention heads per block. The feed-forward sublayers use an expansion factor of $4 \times d$, GELU activations, and dropout with rate 0.1 in all residual paths. Bayesian hyperparameter optimization was performed [Snoek et al., 2012], minimizing validation loss over log-uniform learning rates, uniform dropout, and discrete architectural and training choices including batch size, model width, depth and attention heads. 15% of the training set was reserved as a validation set. Training was performed using the AdamW [Loshchilov and Hutter, 2017] optimizer with learning rate $2 \cdot 10^{-4}$, no weight decay, gradient clipping [Pascanu et al., 2013] at 1.0, micro-batch size 32, and bfloat16 precision.

The computational cost of training scales linearly with the number of glyph-variation examples in the dataset. Let $N$ denote the number of training examples and $P$ the number of control points in a glyph. For each forward pass, the dominant cost arises from self-attention over the point sequence, yielding complexity

$$\mathcal{O}(P^2 d)$$

where $d$ is the latent dimensionality. Consequently, overall training complexity scales as

$$\mathcal{O}(NP^2 d)$$

Table 7: Generated variations for the static UnifrakturMaguntia font under different axis configurations. The text can be selected to verify that it is a real and novel functional variable font rather than a static image.

| Axes | Rendered Text |
|---|---|
| wght<br>slnt<br>wdth<br>opsz | The quick brown fox jumps over the lazy dog |
| wght<br>slnt<br>wdth<br>opsz | The quick brown fox jumps over the lazy dog |
| wght<br>slnt<br>wdth<br>opsz | The quick brown fox jumps over the lazy dog |
| wght<br>slnt<br>wdth<br>opsz | The quick brown fox jumps over the lazy dog |
| wght<br>slnt<br>wdth<br>opsz | The quick brown fox jumps over the lazy dog |
| wght<br>slnt<br>wdth<br>opsz | The quick brown fox jumps over the lazy dog |

In practice, for Roman-script fonts, the number of control points per glyph is moderate. However, CJK glyphs may contain hundreds of control points and occasionally more. In our experiments, to accelerate training and limit memory consumption, we restrict the maximum number of control points per glyph to $P \leq 500$. Glyphs exceeding this threshold remain part of the dataset splits but are excluded during training and evaluation. This bounds the quadratic attention cost and the memory footprint. This restriction is purely practical and can be relaxed or removed to support glyphs with higher numbers of control points.

In variable font generation, the model's average latency is 8.3 ms per tuple on an RTX 3090 GPU. Since tuples can be processed in parallel (subject to the number of glyphs and available memory), the overall font generation process is extremely short.

# G Comparison to Rule-Based Geometric Heuristics

To further examine the limitations of classical axis-specific transformations, we consider the Commissioner variable font, which belongs to our test set and was not observed during training. We constructed a heuristic baseline by applying a global shear transformation to the base glyph and manually tuning the shear angle to approximate the designer-defined ground-truth instance corresponding to $\texttt{slnt} = -1.0$.

We then removed all variability information from the font and applied NIV to predict the slant-axis variation directly from the static outline. Figure 3b presents a qualitative comparison. The ground-truth slanted glyph is shown in green, the heuristic shear-based approximation in red, and the variation predicted by NIV in blue. The prediction produced by NIV more closely matches the ground truth in both structural alignment and local geometric behavior.

This discrepancy arises because slanted or italic variants are not purely the result of a global shear. In practice, they involve a combination of shear, local rotations, stroke-level adjustments, and other non-linear deformations that are difficult to capture using simple analytic transformations. These observations highlight the limitations of naive geometric heuristics and motivate the need for more expressive deformation models.

Table 8: Generated variations for the symbolic typeface Wingdings under different axis configurations. Despite consisting of pictographic symbols rather than alphabetic glyphs, the generated variable font exhibits consistent multi-axis deformation while preserving symbol recognizability.

| Axes | Rendered Text |
|---|---|
| wght<br>slnt<br>wdth<br>opsz | 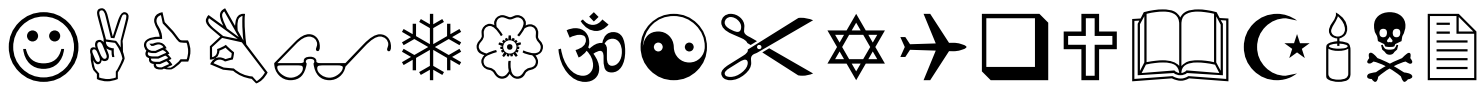 |
| wght<br>slnt<br>wdth<br>opsz | 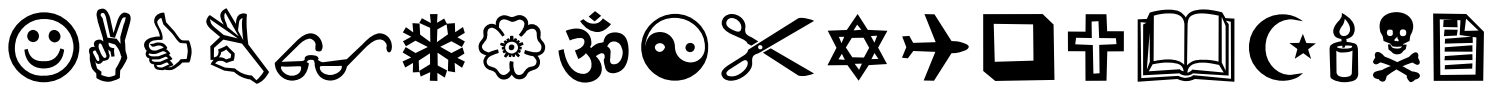 |
| wght<br>slnt<br>wdth<br>opsz | 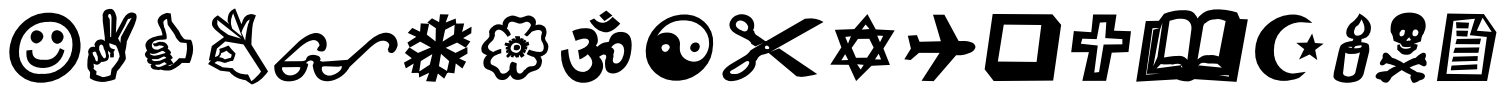 |
| wght<br>slnt<br>wdth<br>opsz | 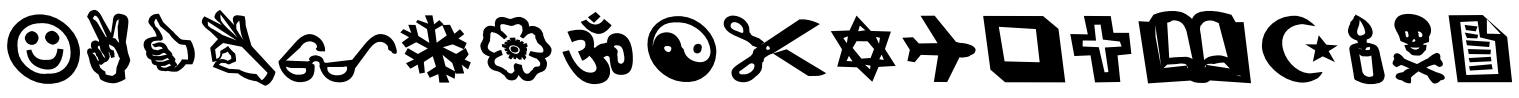 |
| wght<br>slnt<br>wdth<br>opsz | 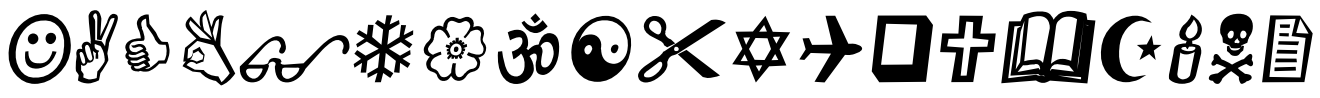 |
| wght<br>slnt<br>wdth<br>opsz | 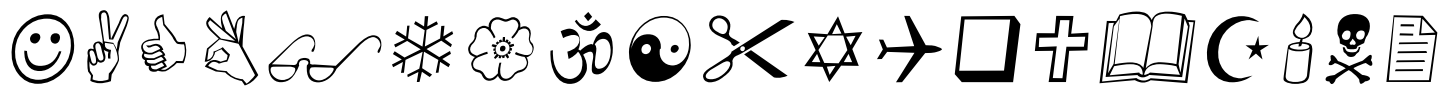 |

Table 9: Handwriting. Rendered pangram using the generated variable font from additional handwriting samples under different axis settings. Try selecting the text to verify that it is a real, usable font and not a static image.

| Axis Configuration | | | | Rendered Text |
|---|---|---|---|---|
| wght | slnt | wdth | opsz | The five boxing wizards jump quickly |
| wght | slnt | wdth | opsz | The five boxing wizards jump quickly |
| wght | slnt | wdth | opsz | The five boxing wizards jump quickly |
| wght | slnt | wdth | opsz | The five boxing wizards jump quickly |
| wght | slnt | wdth | opsz | The five boxing wizards jump quickly |
| wght | slnt | wdth | opsz | The five boxing wizards jump quickly |
| wght | slnt | wdth | opsz | The five boxing wizards jump quickly |
| wght | slnt | wdth | opsz | The five boxing wizards jump quickly |

Table 10: Optical size comparison for “ABNO”.

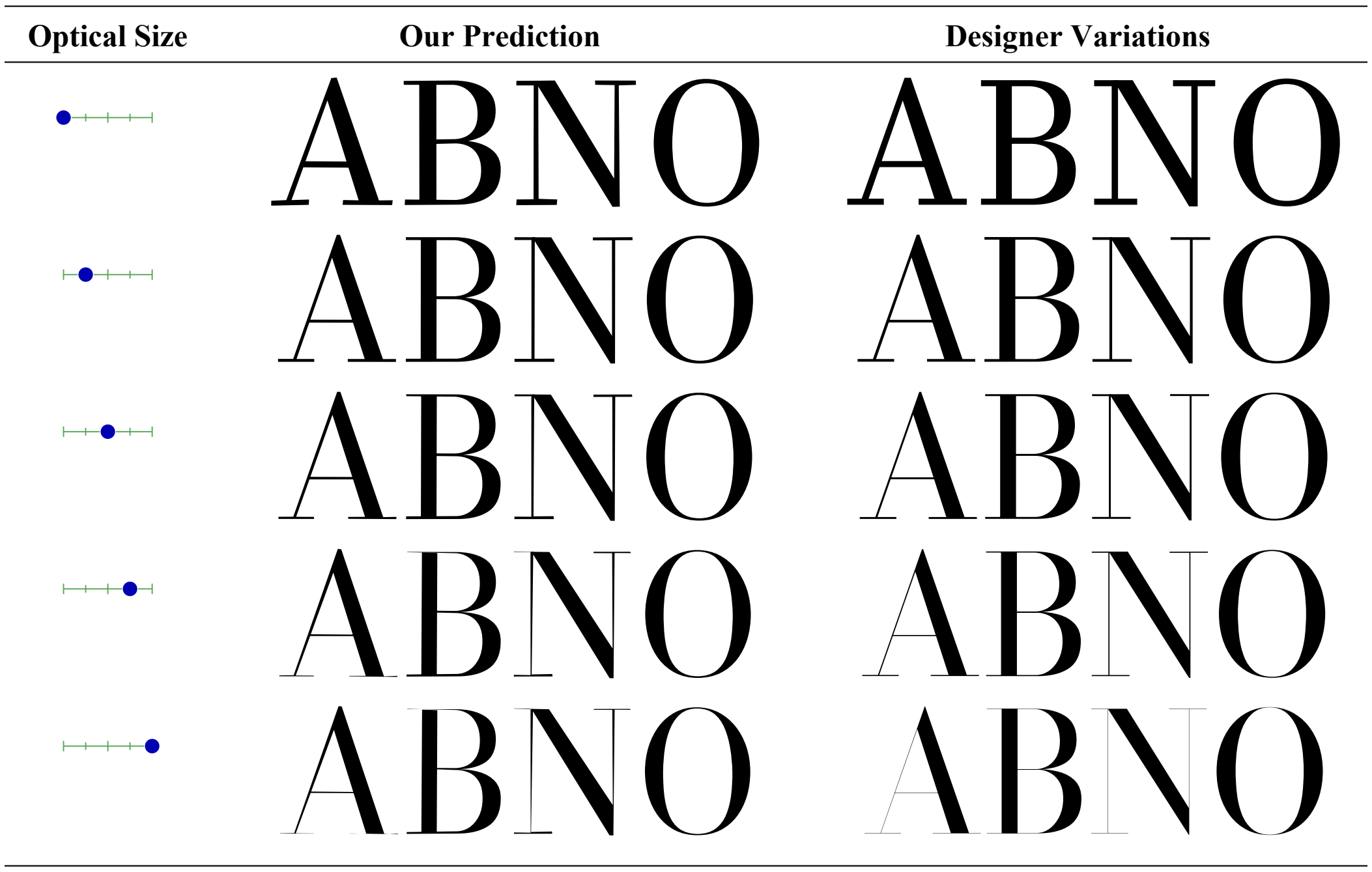

Table 11: Rendered Korean phrase using a novel variable font automatically generated from the static Nanum Myeongjo typeface by NIV.

| Axis Configuration | Rendered Text |
|---|---|
| **Korean (Nanum Myeongjo)** | |
| original | 재빠른 갈색 여우가 게으른 개를 뛰어넘었다 |
| `wght` `slnt` `wdth` | 재빠른 갈색 여우가 게으른 개를 뛰어넘었다 |
| `wght` `slnt` `wdth` | *재빠른 갈색 여우가 게으른 개를 뛰어넘었다* |
| `wght` `slnt` `wdth` | **재빠른 갈색 여우가 게으른 개를 뛰어넘었다** |
| `wght` `slnt` `wdth` | ***재빠른 갈색 여우가 게으른 개를 뛰어넘었다*** |
| `wght` `slnt` `wdth` | 재빠른 갈색 여우가 게으른 개를 뛰어넘었다 |

## H Cross-Axis Interactions are Necessary

Our model is evaluated by predicting an outline deformation (delta on control points) conditioned on the requested axis settings. Concretely, for an input glyph with base control points $\mathbf{P}_0$ and an axis-setting vector $\mathbf{v} \in \mathbb{R}^A$ (sparse, only a few active axes per sample), the model predicts

$$\widehat{\Delta \mathbf{P}} = f_\theta(\mathbf{P}_0, \mathbf{v}) \tag{7}$$

and we report mean squared error (RMSE) between $\widehat{\Delta \mathbf{P}}$ and the ground-truth $\Delta \mathbf{P}$ on the test set.

To test whether the model can decompose multi-axis coordination into independent per-axis contributions, we construct an *additive (no-interaction) evaluation baseline* using the same trained model. For a test sample from the font-split dataset whose active axes are $S$ (e.g., `wght` and `wdth`), let $\mathbf{v}^{(a)}$ be the same axis vector as $\mathbf{v}$ but with only axis $a$ active and all other axes masked (exactly as missing slots are masked during training). We then

Table 12: Implementation details and training hyperparameters.

| **Item** | **Value** |
|---|---|
| Latent dim ($d$) | 256 |
| Layers / heads | 6 layers, 8 heads |
| FF expansion | $4 \times d$ |
| Dropout | 0.1 |
| Positional encoding | sinusoidal |
| Loss | RMSE |
| Epochs | 80 |
| Micro-batch size | 32 |
| Optimizer | AdamW |
| Learning rate | $2 \cdot 10^{-4}$ |
| Weight decay | 0.0 |
| Grad clip | 1.0 |
| Precision | bf16 |
| Hardware | RTX 3090 |
| Peak GPU memory (VRAM) | 4GB |
| Peak system memory (RAM) | 50GB |
| Training wall-clock time | 4 days |
| Total steps | 4.5M |

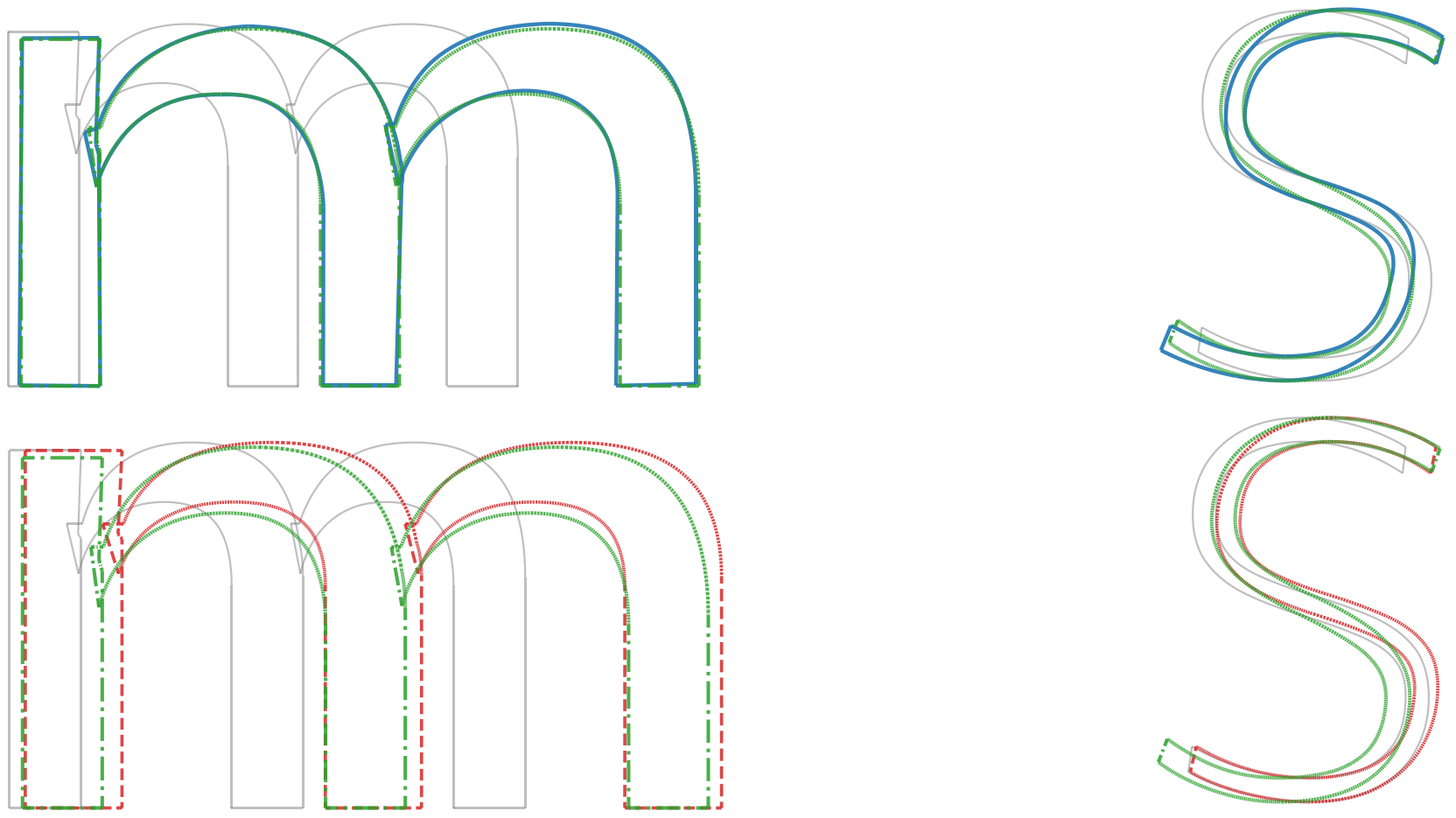

(a) Comparison for the width axis ($\texttt{wdth} = 1.0$) using a heuristic x-scale transformation as baseline. The heuristic increases stem widths, violating width-axis semantics, while NIV preserves stem thickness and adjusts only horizontal proportions.

(b) Comparison for the slant axis ($\texttt{slnt} = -1.0$) using a heuristic shear-based transformation as baseline. The learned model matches the ground truth more closely than the global shear baseline.

Figure 3: Qualitative comparison between NIV (top) and a rule-based heuristic (bottom). Base glyph (gray), ground truth (green), our prediction (blue), and heuristic (red).

run the model once per active axis and sum the predicted deltas:

$$\widehat{\Delta \mathbf{P}}_{\text{add}} = \sum_{a \in S} f_\theta(\mathbf{P}_0, \mathbf{v}^{(a)}) \tag{8}$$

Intuitively, if axis effects composed independently, then (8) should match (7) closely. In practice, this additive evaluation is consistently worse, indicating that meaningful predictions require *cross-axis interactions* inside the model, namely, the joint (nonlinear) processing of multiple axes together.

Table 13: RMSE for additive, per-axis decomposition versus a joint conditioning, on the font-split dataset. For the main method we report mean $\pm$ one standard deviation ($1\sigma$) over 3 runs to indicate statistical stability.

| Evaluation protocol | Test RMSE $\downarrow$ |
|---|---|
| Additive per-axis | 0.05737 |
| Joint axes | $\mathbf{0.05205 \pm 0.00088}$ |

Table 13 demonstrates that the additive decomposition increases RMSE, supporting the conclusion that the model is not simply learning separable axis-specific deformations, but rather relies on axis interactions to predict coordinated geometry.

Designer made variable fonts commonly contain tuples up to order of four axis coordinates in a given tuple coordinates, even when they support more than four axes in the font file. Similarly, when we create automatic variable fonts we recommend to generate tuples up to an order of four. A qualitative evidence of why high order tuples are needed when generating a variable font using our method can be seen in Table 14.

Table 14: The effect of decreasing the maximum tuple coordinate order in a newly created variable font version of the static font Arial. When not allowing four axis interactions in the font tuple generation, the letters 'c' and 'e' are distorted in width, and the weight (boldness) of most characters is not preserved.

| **Max font tuple order** | **Rendered Text** |
|---|---|
| 4 | The five boxing wizards jump quickly |
| 3 | The five boxing wizards jump quickly |
| 2 | The five boxing wizards jump quickly |
| 1 | The five boxing wizards jump quickly |

# I Ablation Study

We run controlled ablation groups on axis embedding strategy and conditioning method. In all runs, we keep the same model/training setup (font split and fixed random seed) and change only the ablated component.

## I.1 Axis Embedding Methods

We compare various methods to build the axis-conditioning signal:

- **Property Embedding (ours)**: the method described in Section 3.1.3.
- **Value MLP**: for each active axis, we sum a learned axis-ID embedding and a value embedding produced by a small MLP (Linear $\rightarrow$ GELU $\rightarrow$ Linear), then aggregate across active axes.
- **Single Embedding**: for each active axis, we scale its axis-ID embedding directly by the scalar axis value, then aggregate across active axes.

### I.2 Conditioning Method

We also ablate how axis embedding is used:

- **Adaptive Layer Normalization, AdaLN (ours)**: in every encoder layer, the two normalization blocks are replaced by adaptive LayerNorm. We first standardize each token across features (LayerNorm-style, without fixed learned affine), then apply condition-dependent scale and shift: $\mathrm{AdaLN}(x, c) = \mathrm{Norm}(x) \odot (1 + \gamma(c)) + \beta(c)$.
- **Feature-wise Linear Modulation, FiLM** [Perez et al., 2018]: encoder layers use a LayerNorm, and we add feature-wise affine modulation after each norm: $h' = \gamma(c) \odot h + \beta(c)$.

Table 15 summarizes all ablations.

Table 15: Ablations on axis embedding and conditioning method. Lower is better. For the main method we report mean $\pm$ one standard deviation ($1\sigma$) over 3 runs to indicate statistical stability.

| Ablation Group | Method | Test RMSE↓ |
|---|---|---|
| Axis Embedding | Value MLP | 0.06644 |
| Axis Embedding | Single Embedding | 0.05405 |
| Conditioning Method | FiLM | 0.05603 |
| **Property Embedding + AdaLN** | | $\mathbf{0.05205 \pm 0.00088}$ |

## J Common Variation Axes in Variable Fonts

Table 16: Generated variations for the original designer-created variable font TikTokSans under single-axis extreme configurations. Each row sets one axis to its extreme while keeping the remaining axes at their default, illustrating isolated axis effects on the rendered text.

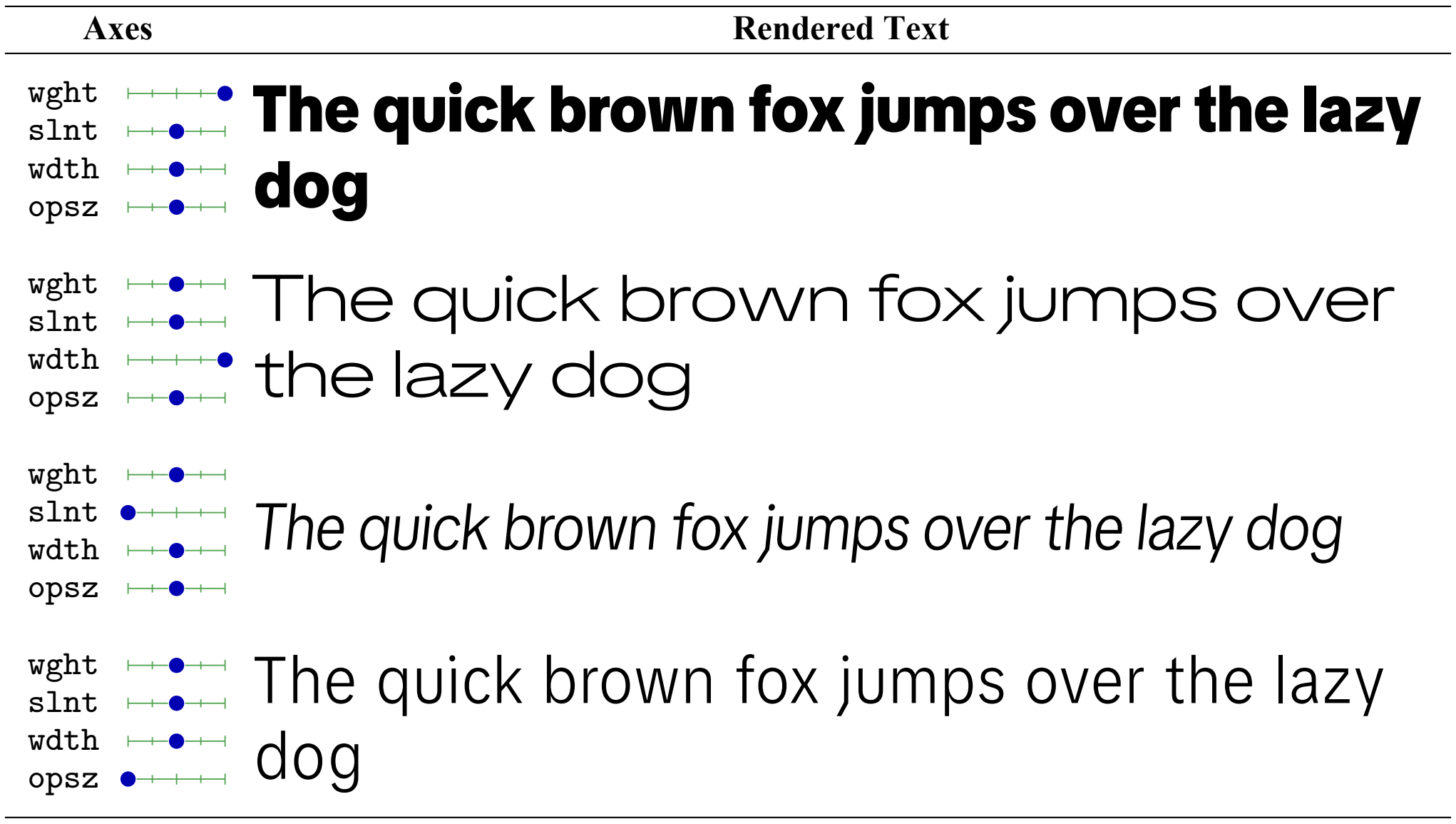

In the OpenType variable font specification, a type designer may define arbitrary custom axes to control glyph variation. These axes can represent any continuous design parameter, such as serif length, contrast, roundness, or other stylistic properties. Despite this flexibility, a small number of axes (Table 16) have become widely adopted and standardized across type families. The most common ones are the following:

**Weight (`wght`).** Controls stroke thickness. Increasing weight makes stems and strokes thicker, producing a transition from light to bold styles, while preserving the overall structure of the glyph.

**Width (`wdth`).** Controls horizontal proportions. Increasing width expands the glyph horizontally without proportionally thickening the strokes. Condensed and extended styles are typically implemented along this axis.

**Slant (`slnt`).** Controls the angle of inclination of the glyph. Unlike true italics, which may introduce structural changes, the slant axis typically applies a geometric inclination of the letterforms.

**Optical Size (`opsz`).** Adjusts the design for different font sizes. Smaller optical sizes usually increase stroke thickness, spacing, and counter openness to improve readability at small text sizes, while larger optical sizes allow finer details and higher contrast for display typography.

Although many other axes exist in practice, these four axes (`wght`, `wdth`, `slnt`, `opsz`) constitute the most common and broadly supported variation dimensions in contemporary variable fonts.

## K Unicode Split

As a further qualitative visualization, in addition to the quantitative evaluation reported earlier, we took the static (non-variable) UnifrakturMaguntia font and the variable New York font, which are not found in either the training or test set. We stripped away all variation information from the variable font, effectively converting it into a static font. We then predicted axis variations for code points in the test set using a model trained with the Unicode-split protocol. Therefore, the model has never seen these specific code points during training, yet it is able to correctly predict the axis variations for them using only the glyph outline information. The results can be seen in Table 17.

Table 17: Rendered New York and UnifrakturMaguntia from the test set code points with axis variations predicted by the Unicode-split-trained model. These specific code points were unseen during training, yet the model correctly predicts their axis variations from glyph outlines alone.

| Axis Configuration | | | | Rendered Code Points |
|---|---|---|---|---|
| **New York** | | | | |
| wght | slnt | wdth | opsz | A D L ` { \| c h t |
| wght | slnt | wdth | opsz | A D L ` { \ c h t |
| wght | slnt | wdth | opsz | A D L ` { / c h t |
| wght | slnt | wdth | opsz | A D L ` { / c h t |
| **UnifrakturMaguntia** | | | | |
| wght | slnt | wdth | opsz | A D L ` { \| c h t |
| wght | slnt | wdth | opsz | A D L ` { \ c h t |
| wght | slnt | wdth | opsz | A D L ` { / c h t |
| wght | slnt | wdth | opsz | A D L ` { / c h t |

## L Public Dataset Construction and Preprocessing Details

To facilitate future research and reproducibility, we fully open-source both the dataset and the preprocessing pipeline used to generate it, enabling researchers to reproduce our results, extend the dataset with additional fonts or axes, and adapt the preprocessing to their own experimental settings. We construct our training dataset from variable TrueType fonts sourced from the Google Fonts collection. Each font is parsed using fontTools, and we extract per-glyph outline representations along with their associated variation deltas from the `gvar` table. To ensure spatial consistency across fonts, which may define their coordinate systems at different resolutions (e.g., 1000, 2000, or 2048 units per em), we normalize all glyph coordinates and variation deltas to a uniform UPM (units per em) of 2048. Composite glyphs, which are constructed by referencing and transforming other glyphs rather than defining their own contours, are excluded, for simplicity. A key step in our preprocessing is the densification of the `gvar` delta representation. The OpenType specification allows variation deltas to be stored sparsely: only a subset of on-curve and off-curve points (the "referenced" points) need explicit delta entries, while the remaining points' deltas are to be inferred at render time according to the specification. This process operates contour-by-contour, inferring each missing point's delta by linearly interpolating between its nearest referenced neighbors along each axis, or clamping to the nearest endpoint delta when the point lies outside the referenced range. We apply this inference during preprocessing so that every glyph–variation tuple contains an explicit delta for every outline point, yielding a fixed-length, fully dense representation suitable for neural network training. Additionally, we retain the four phantom points appended by TrueType (representing the left side bearing, advance width, top side bearing, and advance height), assigning zero deltas where they are absent, so that metric variations are also captured. We further filter variation tuples by axis subset (e.g., retaining only wght, wdth, opsz, and slnt), discarding any tuple that references axes outside the target set. Each glyph's Unicode code point mapping is preserved alongside the geometric data, and the full per-font output is serialized into a structured XML-like format.

As can be seen in Table 18, Google Fonts currently contains 733 variable fonts. Among them, 717 include a `wght` axis, 130 include `wdth`, 42 include `opsz`, and 22 include `slnt`. In addition, there is a long tail of less frequent axes such as `ELSH`, `CRSV`, `ELXP`, and others. Across the collection, there are 1,430,086, 326,744, 108,433, and 38,030 tuples referencing the `wght`, `wdth`, `opsz`, and `slnt` axes, respectively.

Table 18: Google Fonts variable-font dataset statistics. "Axis usage" counts how many times each axis tag appears across all tuple coordinates (hence the Total may exceed the number of tuples).

| **Fonts** | |
|---|---|
| Fonts | 733 |
| **Unicode statistics** | |
| Distinct unicodes | 61,385 |
| Total glyphs | 999,155 |
| **Axis usage across tuple coordinates** | |
| wght | 1,511,137 (66.58%) |
| wdth | 382,328 (16.84%) |
| opsz | 152,651 (6.73%) |
| slnt | 57,023 (2.51%) |
| Others | 166,588 (7.34%) |
| Total | 2,269,727 |
| **Font axis-count distribution (axes per font; `fvar`)** | |
| 1 axis | 541 (73.81%) |
| 2 axes | 155 (21.15%) |
| 3 axes | 10 (1.36%) |
| 4 axes | 13 (1.77%) |
| 5 axes | 7 (0.95%) |
| 11 axes | 6 (0.82%) |
| 13 axes | 1 (0.14%) |
| Total fonts | 733 |
| **Tuple axis-count distribution (axes per tuple)** | |
| 1 axis | 1,288,090 (81.64%) |
| 2 axes | 256,929 (16.28%) |
| 3 axes | 29,563 (1.87%) |
| 4 axes | 3,164 (0.20%) |
| Total tuples | 1,577,746 |

We release the dataset in a human-readable XML format, split into training and test sets (15% test). The total dataset size is 12GB.

### L.1 Dataset Splits and Style Leakage

As with many datasets, perfectly eliminating all possible correlations between the training and test splits is inherently difficult. While our Unicode and font splits are designed to evaluate different generalization properties, some forms of structural or stylistic similarity may still exist across partitions.

**Unicode split.** In the Unicode split, glyphs are separated by code point so that a given Unicode character appears exclusively in either the training or test set. However, Unicode contains cases where different code points correspond to visually identical or nearly identical glyphs (for example compatibility characters or accented variants). While such cases exist, they constitute only a small fraction of the dataset. More commonly, different characters may share strong geometric similarity, such as a base letter and its accented variants (e.g., "a" versus "á"), or characters constructed from similar structural primitives. As a result, some geometric similarity between training and test samples may remain despite the code point separation. Furthermore, Unicode normalization (e.g., NFC — Normalization Form Canonical Composition, and NFKC — Normalization Form Compatibility Composition) would not fully eliminate this effect. While normalization resolves canonical equivalences, many visually similar glyphs remain assigned to distinct Unicode code points. Unicode also includes compatibility and stylistic variants with nearly identical geometry, therefore even normalization or equivalence grouping would not completely remove potential geometric leakage.

**Font split.** In the font split, entire fonts are assigned to either training or test sets. Nevertheless, stylistic similarities may still arise. Fonts may belong to the same family or to larger "super-families" sharing design principles. In addition, some fonts are created by the same type designers or foundries, which may result in recurring stylistic patterns across otherwise distinct fonts. More generally, many fonts share common typographic conventions, which may introduce partial stylistic correlations between splits.

**Practical considerations.** Given these factors, it is difficult to define a perfectly orthogonal partition without introducing substantial complexity or subjective grouping criteria. To keep the dataset construction simple and reproducible, we therefore adopt random partitioning under the Unicode split and font split definitions described earlier. We acknowledge that minor forms of leakage may exist, but emphasize that as long as this work and subsequent works use the same split methodology, results remain directly comparable.

To further demonstrate generalization beyond the training distribution, we also include qualitative experiments on fonts that are completely outside both the training and test sets, including previously non-variable fonts and out-of-distribution handwriting fonts. These examples provide additional evidence of the model's ability to generalize beyond memorized geometry.

## M Dataset License

The dataset released with this work is constructed from variable fonts obtained from the Google Fonts collection. We do not modify or redistribute altered versions of the original font binaries.

Each font included in the dataset remains subject to its original license as provided by its respective copyright holder. The dataset preserves the original licensing terms of each font, and users are responsible for complying with those terms when using the data.

The preprocessing scripts and metadata introduced in this work are released separately under an open-source license.

## N Societal Impact

On the positive side, the method lowers the technical and financial barriers to creating high-quality variable fonts. This may democratize typography, enabling independent designers, small studios, and underrepresented language communities to produce expressive and functional fonts without extensive manual engineering. It can also support accessibility, for example by enabling dynamic adjustment of weight or width to improve readability for visually impaired users. In addition, automating repetitive font-engineering steps may reduce labor-intensive workflows and increase productivity in digital publishing and user interface design.

On the negative side, automation may reduce demand for certain specialized manual font-engineering tasks, potentially affecting professional type designers.

Overall, the societal impact is moderate and primarily related to creative industries and digital communication.

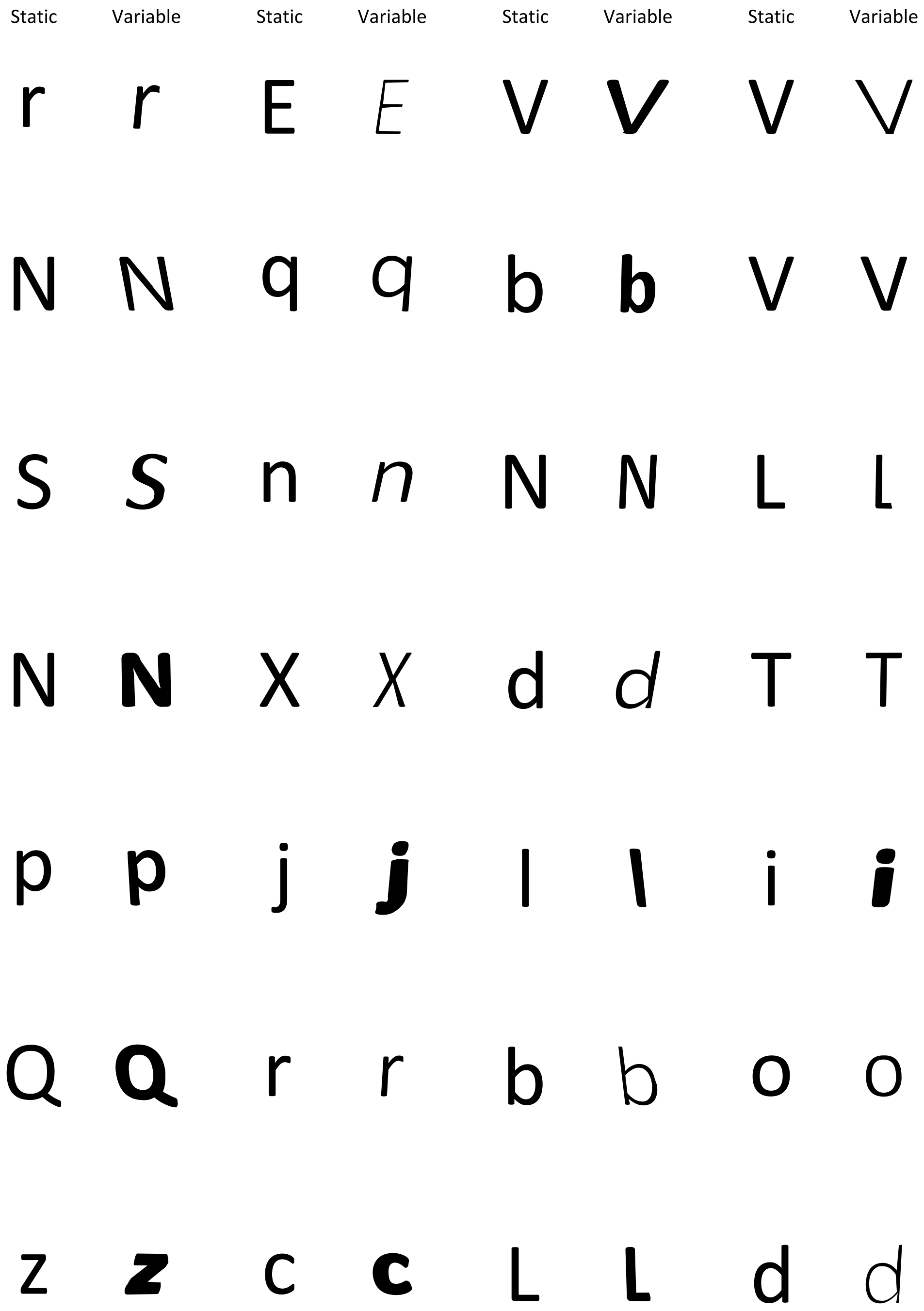

Figure 4: Results of random axis variations on Calibri and failure cases. A qualitative comparison using the same variable font in both columns. In each cell, the left glyph uses the base configuration with all four axes set to zero, and the right glyph uses a random four-axis setting over `wght`, `wdth`, `slnt`, and `opsz`. Each pair shows one alphabetic letter.